\documentclass[conference]{IEEEtran}
\IEEEoverridecommandlockouts
\usepackage[compatibility=false]{caption}
\usepackage{cite}
\usepackage{amsmath,amssymb,amsfonts}
\usepackage{algorithmic}
\usepackage{graphicx}
\usepackage{textcomp}
\usepackage{xcolor}
\def\BibTeX{{\rm B\kern-.05em{\sc i\kern-.025em b}\kern-.08em
    T\kern-.1667em\lower.7ex\hbox{E}\kern-.125emX}}
\usepackage{graphicx}
\usepackage{subfig}
\usepackage{multirow}
\usepackage{amsmath}
\usepackage{cleveref}

\begin{document}

\title{JSSFF: A Joint Structural–Semantic Fusion Framework for Remote Sensing Image Captioning\\
}

\author{\IEEEauthorblockN{1\textsuperscript{st} Swadhin Das}
\IEEEauthorblockA{\textit{Computer Science and Engineering} \\
\textit{University of Petroleum and Energy Studies}\\
Dehradun, India \\
swadhin.das@ddn.upes.ac.in}
\and
\IEEEauthorblockN{2\textsuperscript{nd} Vivek Yadav}
\IEEEauthorblockA{\textit{Comuputer Science and Engineering} \\
\textit{University of Petroleum and Energy Studies}\\
Dehradun, India\\
vivek.yadav@ddn.upes.ac.in}
% \and
% \and
% \and
% \and
}

\maketitle

\begin{abstract}
The encoder--decoder framework has become widely popular nowadays. In this model, the encoder extracts informative visual features from an input image, and the decoder employs a sequence-to-sequence formulation to generate the corresponding textual description from these features. The existing models focus more on the decision part. However, extracting meaningful information from the image can help the decoder generate an accurate caption by providing information about the objects and their relationship. Remote sensing images are highly complex. One major challenge is detecting objects that extend beyond their visible boundaries due to occlusion, overlapping structures, and unclear edges. Hence, there is a need to design an approach that can effectively capture both high-level semantics and low-level spatial details for accurate caption generation. In this work, we have proposed an edge-aware fusion method by incorporating the original image and its edge-aware version into the encoder to enhance feature representation and boundary awareness. We used a comparison-based beam search (CBBS) to generate captions to achieve a balanced trade-off between quantitative metrics and qualitative caption relevance through fairness-based comparison of candidate captions. Experimental results demonstrate our model's superiority over several baseline models in quantitative and qualitative perspectives.
\end{abstract}

\section{Introduction}
Encoder--decoder-based captioning frameworks are the most widely adopted approaches for remote sensing image captioning (RSIC)~\cite{qu2016deep,hoxha2020new}. In these approaches, the encoder is used to extract some informative visual representations (features) from the image. The decoder then produces a word sequence based on this representation. CNN backbones (such as ConvNeXt, ResNet) are commonly used as encoders~\cite{das2024unveiling}, while LSTM~\cite{das2024unveiling,das2024textgcn}, GRU~\cite{das2025fe}, or even SVM-based decoders~\cite{hoxha2021novel} have been explored for caption generation.

Remote sensing images, sourced from satellites, aircraft, and UAVs, exhibit non-standard viewpoints, high scene density, and large variations in scale, orientation, and spatial resolution~\cite{lu2017exploring}. These characteristics make it difficult to extract both fine-grained object cues and their spatial relationships. A single encoder often fails to capture this visual complexity~\cite{das2024unveiling}. Das et al.~\cite{das2025fe} introduced a Multi-stream Encoder–Decoder Framework (MsEdF) that fuses representations from two separate encoders to mitigate the limitations of the single encoder-based framework. However, the design still underutilized spatial structure and low-level cues such as edges, limiting its ability to model geometric properties within remote sensing scenes.

Another major issue in the existing RSIC models~\cite{das2020correcting,hoxha2021novel,ren2022mask} is the lack of focus on finding intricate spatial patterns and low-level visual features (such as edges). Das et al.,~[2025]~\cite{das2025novel} have proposed an edge-aware technique that incorporates enhancing the spatial patterns, thus improving the images for the encoder to capture the low-level features. However, this work primarily focused on improving the extraction of low-level features without leveraging the complementary relationship between the original and edge-aware representations. 

In our work, we address these limitations by fusing both the original and edge-aware images through an early fusion technique, enabling the encoder to jointly learn structural and semantic cues for a more robust feature representation. The contribution of this work is as follows.
\begin{itemize}
    \item An early fusion-based encoding technique is employed to extract features from two different versions of the input image to enhance spatial correspondence and improve feature representation for boundary-sensitive regions.
    \item An edge-aware representation is utilized to incorporate fine-grained structural and low-level visual cues, enabling better object localization and boundary detection in complex remote sensing scenes.
    \item A comparison-based beam search (CBBS) mechanism is adopted during caption generation to achieve a balanced trade-off between quantitative performance and qualitative caption relevance.
    \item Extensive quantitative and qualitative experiments on multiple benchmark datasets validate the effectiveness of the proposed model.
\end{itemize}

The rest of the work is as follows. In~\Cref{sec:related}, we present a review of the literature relevant to our work.~\Cref{sec:proposed} described the proposed method with novel components. The experimental setup used in our model is discussed in~\Cref{sec:exp_setup}. The justification of our work using both quantitative and qualitative methods is presented in~\Cref{sec:exp_results}. Finally, we have concluded our work in~\Cref{sec:conclusion}.
\section{Related Work}
\label{sec:related}
In the encoder--decoder framework, a CNN is typically used to extract visual features from an image, while an RNN generates the corresponding textual description of these features in RSIC~\cite{qu2016deep,das2025fe}. Lu et al.,~[2017]~\cite{lu2017exploring} proposed a new dataset \emph{RSICD} for RSIC. Xu et al.~[2017]~\cite{xu2017image} proposed a deep residual LSTM network with identity mappings and a novel temporal dropout strategy to improve training stability and reduce overfitting in image caption generation. Zhang et al.,~[2019]~\cite{zhang2019multi} proposed a multiscale crooping mechanism to extract visual features from different dimensions. Das et al.,~[2020]~\cite{das2020pso} proposed a PSO-based gamma correction method to enhance the visual presentation of an image. Hoxha et al.,~[2020]~\cite{hoxha2020new} proposed a new search technique, called Comparison-based Beam Search (CBBS), to improve the effectiveness of traditional beam search. Hoxha et al.,~[2021]~\cite{hoxha2021novel} proposed a novel SVM-based decoding strategy for RSIC. Sairam et al.,~[2021]~\cite{sairam2021image} developed a CNN-LSTM-based encoder–decoder framework that generates multilingual image captions. Ye et al.~[2022]~\cite{ye2022joint} proposed a two-stage captioning method of joint training using multilabel classification. Hoxha et al.,~[2023]~\cite{hoxha2023improving} increased caption precision by optimizing post‑processing of predicted word sequences.

Several attention mechanisms have been explored in encoder--decoder architectures to overcome CNN-LSTM's limitation of uniform focus on the entire image. By selectively attending to informative regions, attention mechanisms enhance fine-grained feature extraction and improve semantic alignment between visual and linguistic representations. Chen et al.~\cite{chen2018show} developed an attribute-driven attention model using an attribute-inference technique to improve caption generation. Li et al.~\cite{li2020multi} applied a multilevel analysis model with three attention structures that occur at the feature level. Wang et al.~\cite{wang2022multi} enhanced captioning by multilevel fusion of spatial and linguistic characteristics. Zia et al.~\cite{zia2022transforming} introduced a multi-scale feature processing technique through an adaptive attention-based decoder.

Although existing methods can generate meaningful descriptions for remote sensing images, they still have several limitations. Most approaches mainly focus on improving the decoder while paying limited attention to the encoder, which restricts the quality of the visual features extracted. These models typically rely on global image features, capturing only the overall information of the scene, but not caring for important local or fine-grained details. As a result, the generated captions often miss small but semantically significant objects or spatial relationships. In this work, we address these issues by enhancing the encoder to better extract both global and local visual features, enabling richer contextual understanding and more accurate alignment between image regions and textual descriptions.

\section{Proposed Method}
\label{sec:proposed}
\begin{figure}[!ht]
    \centering
    \includegraphics[width=\linewidth,height=150px]{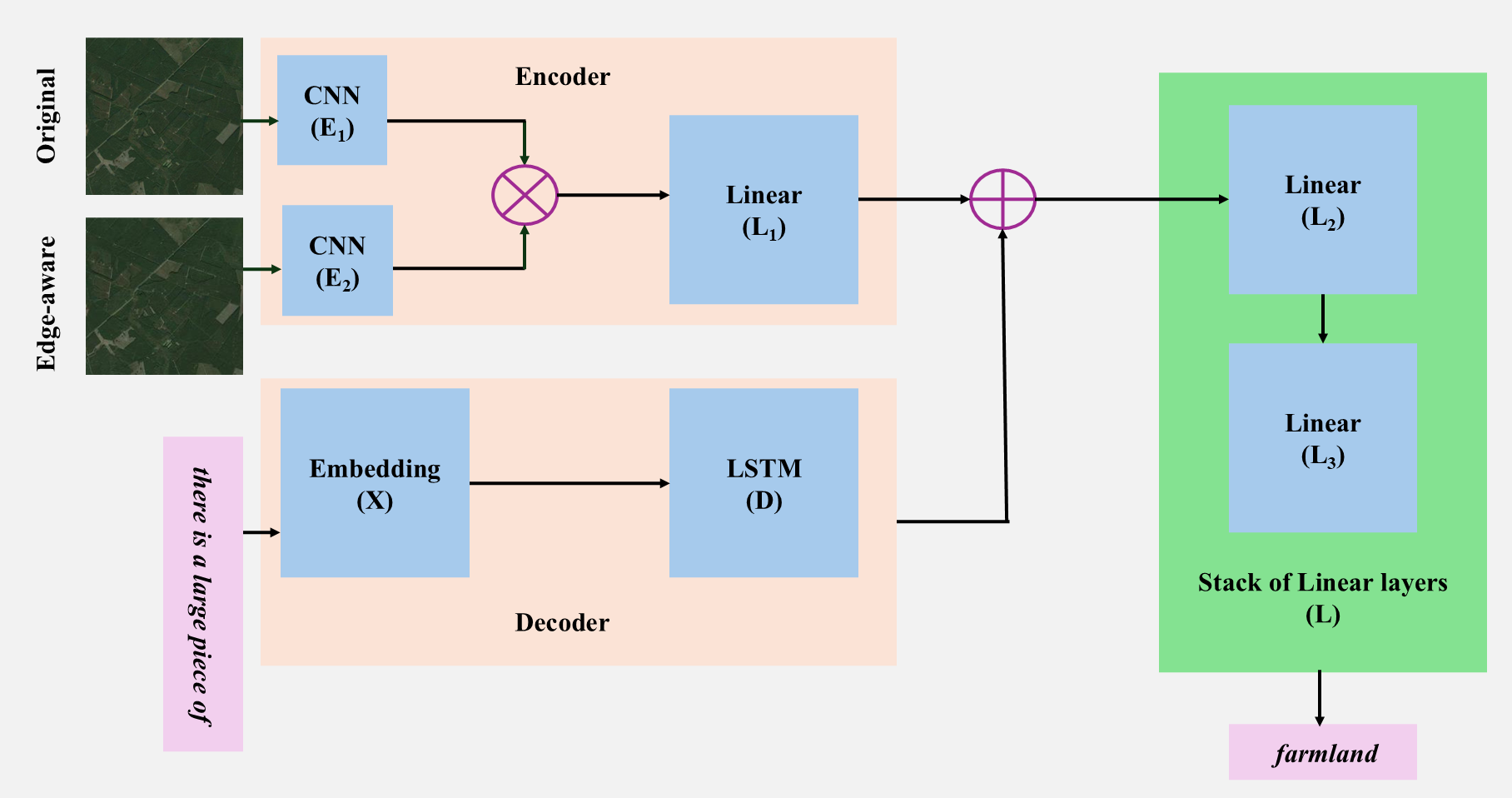}
    \caption{Architecture of the Proposed Model}
    \label{fig:architecture}
\end{figure}
We use a standard encoder–decoder architecture. Two inputs are provided to the CNNs ($E_1$ and $E_2$): the original image and its edge-aware version. ConvNext~\cite{liu2022convnet} serves as the backbone, as it is one of the most effective CNNs for remote sensing applications~\cite{das2025good}. The outputs of the two branches are then fused using position-wise concatenation, preserving spatial correspondence before fusion. The fused features are passed to a linear layer ($L_1$). For the decoder, we adopt a sequence-to-sequence model in which the input sequence is embedded via ($X$) and then processed by the decoder ($D$) to model temporal dependencies (we use LSTM). The outputs of $L_1$ and $D$ are concatenated and sent to a \emph{stack of linear layers} ($L$), which produces the vocabulary-level probability distribution for the next token. An overview of the method is shown in~\Cref{fig:architecture}, where $\otimes$ denotes position-wise concatenation and $\oplus$ denotes conventional concatenation.

Let two sequences (or vectors) be defined as
\[
P = [p_0, p_1, p_2, \ldots, p_n]
\quad \text{and} \quad
Q = [q_0, q_1, q_2, \ldots, q_n]
\]

the position-wise concatenation operation is defined as
\[
P \otimes Q = [p_0, q_0, p_1, q_1, p_2, q_2, \ldots, p_n, q_n]
\]

and the conventional concatenation is defined as
\[
P \oplus Q = [p_0, p_1, p_2,\ldots, p_n, q_0, q_1, q_2,\ldots, q_n]
\]
\subsection{Edge-aware Image Representation}
\begin{figure}[!ht]
    \centering
    \includegraphics[height=105px,width=\linewidth]{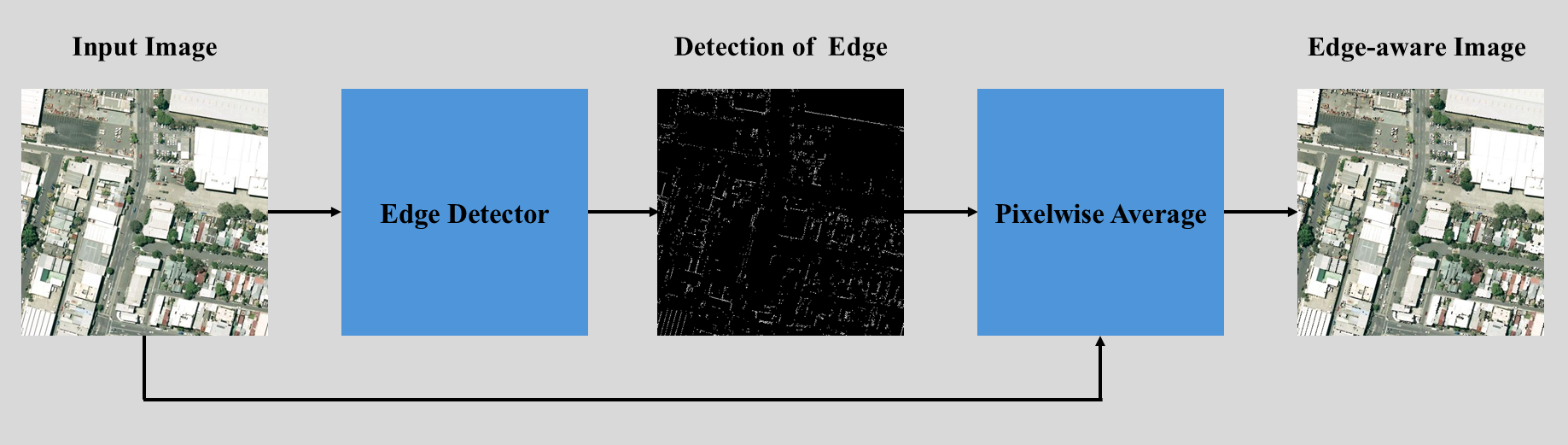}
    \caption{Architecture of the Fusion Mechanism with the Edge-Aware Module}
    \label{fig:edge_aware}
\end{figure}
Edge-aware images are effective in extracting meaningful information by enabling accurate object detection and identifying similarities among objects while preserving spatial information from the image~\cite{das2025novel}. In this work, we used the edge-aware version along with the original image to extract two different image representations rather than obtaining complementary representations of the same image through different encoders. The process of generating the edge-aware image from the original image is illustrated in \Cref{fig:edge_aware}. In this work, we have used three commonly used edge detectors: Canny~\cite{canny1986computational}, Sobel~\cite{sobel19683x3}, and Laplacian~\cite{marr1980theory} (similar to~\cite{das2025novel}).
\begin{table}[!ht]
\caption{Comparison of Different Edge Detectors on the SYDNEY Dataset}
\label{SYDNEY_edge}
\resizebox{\linewidth}{!}{
\begin{tabular}{|c|c|c|c|c|c|c|c|}
    \hline
    Edge & BLEU-1 & BLEU-2 & BLEU-3 & BLEU-4 & METEOR & ROUGE-L & CIDEr \\
    \hline
    Original & 0.8104 & 0.7158 & 0.6376 & 0.5682 & 0.4355 & 0.7540 & 2.7014 \\
    \hline
    Canny & 0.8095 & 0.7226 & 0.6445 & 0.5741 & 0.4336 & 0.7486 & 2.6569 \\
    \hline
    Sobel & 0.8148 & 0.7274 & \textbf{0.6526} & \textbf{0.5851} & 0.4402 & 0.7568 & 2.6783 \\
    \hline
    Laplacian & \textbf{0.8215} & \textbf{0.7299} & 0.6502 & 0.5806 & \textbf{0.4498} & \textbf{0.7664} & \textbf{2.7168} \\
    \hline
\end{tabular}}
\end{table}
\begin{table}[!ht]
\caption{Comparison of Different Edge Detectors on the UCM Dataset}
\label{UCM_edge}
\resizebox{\linewidth}{!}{
\begin{tabular}{|c|c|c|c|c|c|c|c|}
    \hline
    Edge & BLEU-1 & BLEU-2 & BLEU-3 & BLEU-4 & METEOR & ROUGE-L & CIDEr \\
    \hline
    Original & 0.8466 & 0.7790 & 0.7277 & 0.7020 & 0.4811 & 0.8121 & 3.4423 \\
    \hline
    Canny & 0.8600 & 0.7964 & 0.7462 & 0.7201 & 0.5048 & 0.8308 & 3.6031 \\
    \hline
    Sobel & 0.8610 & 0.8001 & 0.7482 & 0.7228 & 0.5027 & 0.8225 & 3.7357 \\
    \hline
    Laplacian & \textbf{0.8717} & \textbf{0.8068} & \textbf{0.7521} & \textbf{0.7299} & \textbf{0.5125} & \textbf{0.8416} & \textbf{3.7812} \\
    \hline
\end{tabular}}
\end{table}
\begin{table}[!ht]
\caption{Comparison of Different Edge Detectors on the RSICD Dataset}
\label{RSICD_edge}
\resizebox{\linewidth}{!}{
\begin{tabular}{|c|c|c|c|c|c|c|c|}
    \hline
    Edge & BLEU-1 & BLEU-2 & BLEU-3 & BLEU-4 & METEOR & ROUGE-L & CIDEr \\
    \hline
    Original & 0.6409 & 0.4678 & 0.3613 & 0.2889 & 0.2547 & 0.4816 & 0.7917 \\
    \hline
    Canny & 0.6468 & 0.4726 & 0.3652 & 0.2919 & 0.2585 & 0.4834 & 0.8093 \\
    \hline
    Sobel & 0.6448 & 0.4742 & 0.3685 & 0.2981 & \textbf{0.2698} & 0.4847 & \textbf{0.8254} \\
    \hline
    Laplacian & \textbf{0.6499} & \textbf{0.4760} & \textbf{0.3724} & \textbf{0.3088} & 0.2624 & \textbf{0.4922} & 0.8159 \\
    \hline
\end{tabular}}
\end{table}

The comparison between two different fusion techniques is presented in~\Cref{SYDNEY_fusion,UCM_fusion,RSICD_fusion}. To independently assess the effect of edge information, we used a single image in each case without applying any fusion. The results clearly demonstrate that edge-aware images effectively capture spatial patterns and low-level visual cues, thereby enhancing caption generation.
\subsection{Fusion Technique}
\begin{figure}[!ht]
  \centering
  \subfloat[Early Fusion\label{fig:early_fusion}]{\includegraphics[width=\linewidth,height=70px]{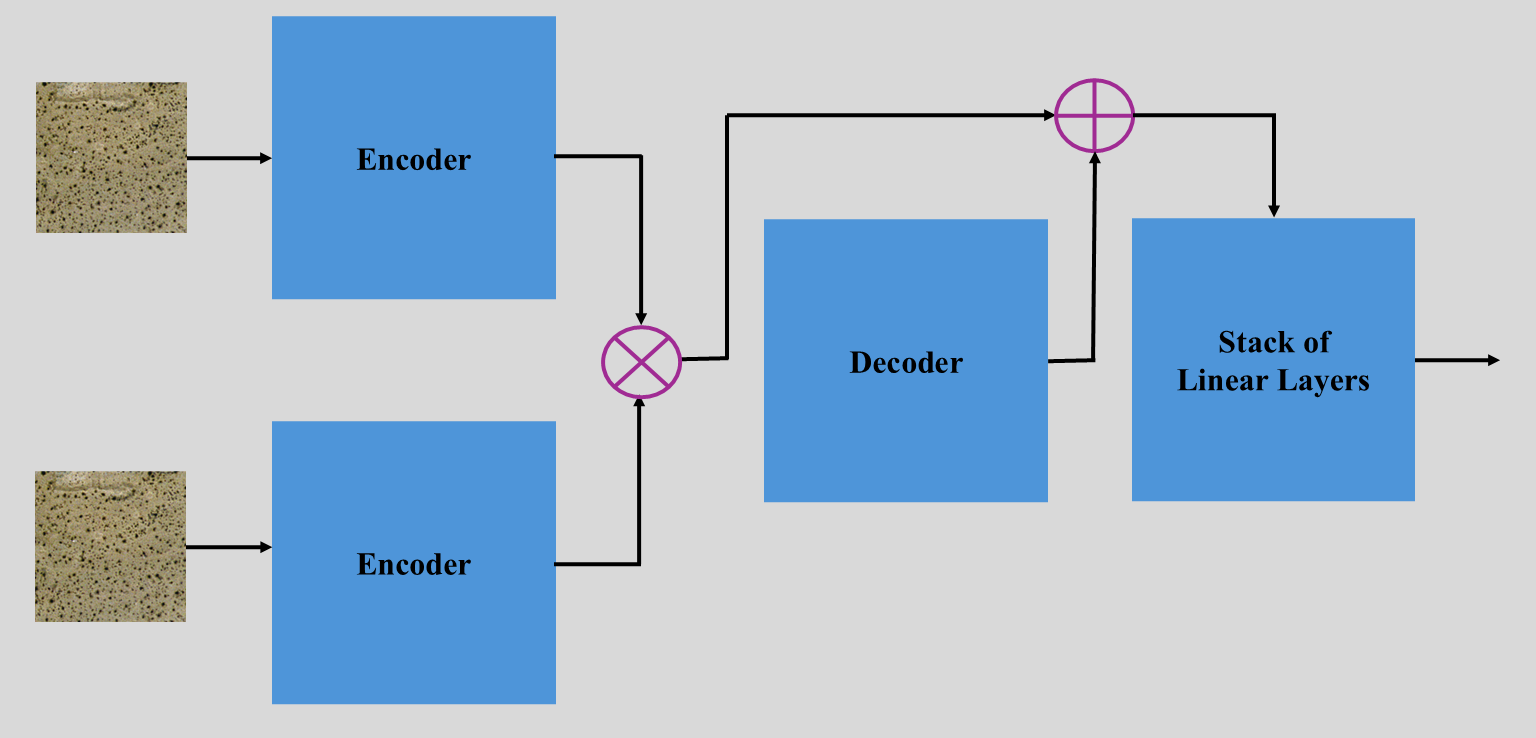}}
  
  \subfloat[Late Fusion\label{fig:late_fusion}]{\includegraphics[width=\linewidth,height=70px]{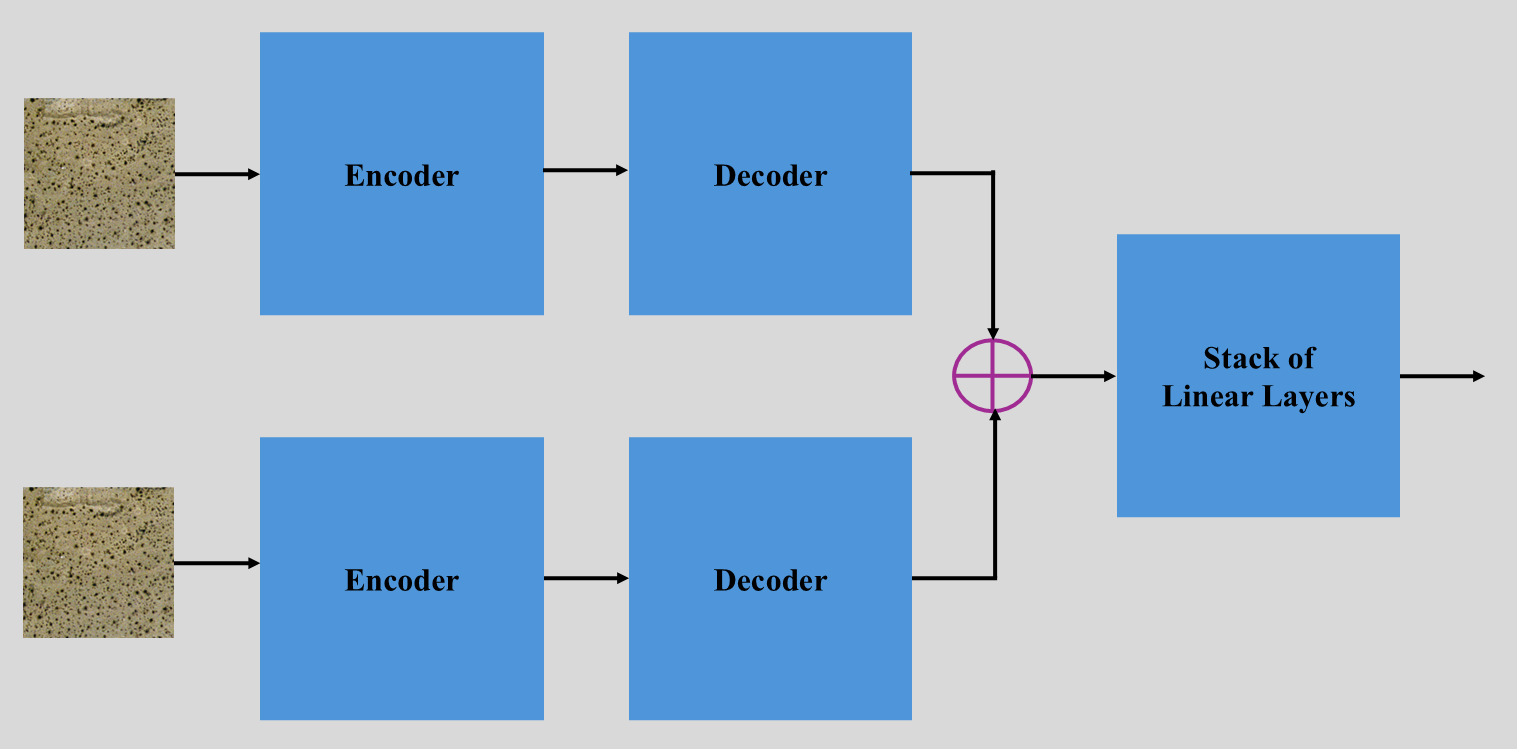}}
   \caption{Architecture of Different Fusion Techniques}
   \label{fig:fusion}
\end{figure}
To effectively combine complementary information from multiple image representations, a suitable fusion technique is essential~\cite{das2025fe}. In this work, we have demonstrated two different types of fusion technique, early fusion and late fusion. \Cref{fig:fusion} illustrate the architecture of these two fusion techniques. In early fusion, we have concatenated the output from the encoders and passed it to a single decoder. In late fusion, a different encoder--decoder setup is used to process two images. Then the result of these two setups is concatenated and passed to \emph{stack of linear layer}.
\begin{table}[!ht]
\caption{Analyzing Different Fusion Techniques on the SYDNEY Dataset}
\label{SYDNEY_fusion}
\centering
\resizebox{\linewidth}{!}{
\begin{tabular}{|c|c|c|c|c|c|c|c|c|}
    \hline
    Image Type & Fusion Type & BLEU-1 & BLEU-2 & BLEU-3 & BLEU-4 & METEOR & ROUGE-L & CIDEr \\
    \hline
    \multirow{2}{*}{Original$\otimes$Canny} & Early & 0.8308 & 0.7518 & 0.6812 & 0.6146 & 0.4589 & 0.7731 & 2.8880 \\
    \cline{2-9}
     & Late  & 0.8111 & 0.7182 & 0.6431 & 0.5781 & 0.4312 & 0.7594 & 2.6891 \\
    \hline
    \multirow{2}{*}{Original$\otimes$Sobel} & Early & 0.8257 & 0.7532 & 0.6875 & 0.6294 & 0.4592 & 0.7849 & 3.0092 \\
    \cline{2-9}
     & Late  & 0.8052 & 0.7178 & 0.6469 & 0.5844 & 0.4369 & 0.7612 & 2.6776 \\
    \hline
    \multirow{2}{*}{Original$\otimes$Laplacian} & Early & \textbf{0.8402} & \textbf{0.7738} & \textbf{0.7196} & \textbf{0.6719} & \textbf{0.4730} & \textbf{0.7909} & \textbf{3.0766} \\
    \cline{2-9}
     & Late  & 0.8128 & 0.7206 & 0.6460 & 0.5832 & 0.4328 & 0.7525 & 2.6332 \\
    \hline
\end{tabular}}
\end{table}
\begin{table}[!ht]
\caption{Analyzing Different Fusion Techniques on the UCM Dataset}
\label{UCM_fusion}
\centering
\resizebox{\linewidth}{!}{
\begin{tabular}{|c|c|c|c|c|c|c|c|c|}
    \hline
    Image Type & Fusion Type & BLEU-1 & BLEU-2 & BLEU-3 & BLEU-4 & METEOR & ROUGE-L & CIDEr \\
    \hline
    \multirow{2}{*}{Original$\otimes$Canny} & Early & 0.8736 & 0.8041 & 0.7451 & 0.7107 & 0.5064 & 0.8322 & 3.7312 \\
    \cline{2-9}
     & Late & 0.8745 & 0.8068 & 0.7526 & 0.7279 & 0.5161 & 0.8330 & 3.4415 \\
    \hline
    \multirow{2}{*}{Original$\otimes$Sobel} & Early & 0.8815 & 0.8138 & 0.7562 & 0.7227 & 0.4984 & 0.8306 & 3.6369 \\
    \cline{2-9}
     & Late & 0.8645 & 0.7926 & 0.7354 & 0.7040 & 0.4982 & 0.8176 & 3.3352 \\
    \hline
    \multirow{2}{*}{Original$\otimes$Laplacian} & Early & \textbf{0.8978} & \textbf{0.8283} & \textbf{0.7715} & 0.7396 & \textbf{0.5296} & \textbf{0.8592} & \textbf{3.8932} \\
    \cline{2-9}
     & Late & 0.8775 & 0.8171 & 0.7685 & \textbf{0.7434} & 0.5241 & 0.8462 & 3.6067 \\
    \hline
\end{tabular}}
\end{table}
\begin{table}[!ht]
\caption{Analyzing Different Fusion Techniques on the RSICD Dataset}
\label{RSICD_fusion}
\centering
\resizebox{\linewidth}{!}{
\begin{tabular}{|c|c|c|c|c|c|c|c|c|}
    \hline
    Image Type & Fusion Type & BLEU-1 & BLEU-2 & BLEU-3 & BLEU-4 & METEOR & ROUGE-L & CIDEr \\
    \hline
    \multirow{2}{*}{Original$\otimes$Canny} & Early & 0.6518 & 0.4838 & 0.3802 & 0.3090 & 0.2704 & 0.4951 & 0.8608 \\
    \cline{2-9}
     & Late & 0.6485 & 0.4786 & 0.3724 & 0.3001 & 0.2637 & 0.4927 & 0.8382 \\
    \hline
    \multirow{2}{*}{Original$\otimes$Sobel} & Early & 0.6564 & 0.4805 & 0.3751 & 0.3025 & 0.2705 & 0.4916 & 0.8416 \\
    \cline{2-9}
     & Late & 0.6484 & 0.4726 & 0.3660 & 0.2944 & 0.2619 & 0.4865 & 0.8236 \\
    \hline
    \multirow{2}{*}{Original$\otimes$Laplacian} & Early & \textbf{0.6630} & \textbf{0.4904} & \textbf{0.3883} & \textbf{0.3204} & \textbf{0.2795} & \textbf{0.5144} & \textbf{0.8842} \\
    \cline{2-9}
     & Late & 0.6527 & 0.4794 & 0.3819 & 0.3098 & 0.2678 & 0.4973 & 0.8298 \\
    \hline
\end{tabular}}
\end{table}

\Cref{SYDNEY_fusion,UCM_fusion,RSICD_fusion} shows the comparison between early and late fusion techniques. From these numerical analysis, it is clearly evident that early fusion outperforms late fusion by significant margins. The main reason behind this result is the joint learning of complementary features at an early stage. Consequently, the model can capture both structural and semantic cues more effectively. In early fusion, we combine the outputs of two different image representations which enriches the image understanding. In contrast, in late fusion the encoder–decoder setup is trained independently for each stream, leading to inconsistent feature alignment and causes unnecessary confusion for \emph{stack of linear layers}, as two separate decision paths are merged at a later stage. Hence, in this work, we have adopted the early fusion model for subsequent experiments.
\subsection{Comparison-based Beam Search}
In sequence-to-sequence caption generation, search strategies play a crucial role in predicting successive words in the output sequence. Traditional methods, such as greedy search~\cite{sutskever2014seq2seq} and beam search~\cite{bahdanau2014neural}, rely solely on the likelihood of each word in the output probability distribution, which can limit the semantic richness, fluency, and contextual relevance of generated captions. To address this, Comparison-based Beam Search (CBBS)~\cite{hoxha2020new,das2024textgcn} extends the standard beam search by incorporating content-level comparisons. CBBS uses beam search as the backbone to generate multiple captions. Then, using the K-Nearest Neighbors (KNN) algorithm, it retrieves visually similar images from a precomputed archive. The captions of these images are considered reference captions. Finally, every candidate caption is compared with the reference captions using a sentence-scoring technique, and the best caption is selected by the highest similarity score. This approach improves the semantic relevance and contextual accuracy of the generated caption. However, it also adds computational overhead and depends on the diversity of the reference archive, potentially reducing performance on novel or rare images.
\section{Experimental Setup}
\label{sec:exp_setup}
In our work, we employ a standard encoder–decoder architecture (refer to \Cref{fig:architecture}). The dimensions of both the embedding layer ($X$) and the LSTM decoder ($D$) are set to $256$. Dimensional consistency with the decoder output is ensured by configuring the first linear layer ($L_1$) to produce a $256$-dimensional representation. Consequently, the second linear layer ($L_2$) takes a $512$-dimensional input and projects it to a $256$-dimensional output. Finally, the third linear layer ($L_3$) projects the learned representation into the vocabulary space of the data set, generating a probability distribution over the next candidate tokens in the sequence. Based on the analysis shown in \Cref{SYDNEY_edge,UCM_edge,RSICD_edge,SYDNEY_fusion,UCM_fusion,RSICD_fusion}, we found that the Laplacian edge detection technique achieved the best performance, consistent with previous findings in transformer-based architectures~\cite{das2025novel}. Hence, we adopt it for further experiments and analysis. Grammarly and ChatGPT were used to refine the manuscript for clarity and grammar without adding any new content.

Three widely used datasets were used to validate the experimental results: SYDNEY~\cite{qu2016deep} ($613$ images), UCM~\cite{qu2016deep} ($2{,}100$ images) and RSICD~\cite{lu2017exploring} ($10{,}921$ images). In our work, we utilized the refined versions of these datasets proposed by~\cite{das2024textgcn}, which deal with several errors present in the original versions. We also followed the same train–val-test splits as provided in the respective datasets. We evaluated our approach using four standard RSIC metrics: BLEU~\cite{papineni-etal-2002-bleu}, METEOR~\cite{lavie-agarwal-2007-meteor}, ROUGE~\cite{lin-2004-ROUGE}, and CIDEr~\cite{vedantam2015cider}.
\section{Experimental Results}
\label{sec:exp_results}
We perform an extensive evaluation of our model using both quantitative and qualitative analyzes. In the quantitative study, we compare our approach with several state-of-the-art RSIC methods. Our qualitative analysis includes subjective evaluation and visual examples from test datasets. Finally, we have conducted an error analysis of these visual results.
\subsection{Comparison of Different RSIC Models}
\begin{table}[!ht]
    \centering
    \caption{Evaluation of Different Remote Sensing Image Captioning Methods on the SYDNEY Dataset}
    \label{SYDNEY_SEARCH}
    \resizebox{\linewidth}{!}{
    {\small
    \begin{tabular}{|c|c|c|c|c|c|c|c|}
    \hline
    METHOD & BLEU-1 & BLEU-2 & BLEU-3 & BLEU-4 & METEOR & ROUGE-L & CIDEr \\
    \hline
    R-BOW \cite{lu2017exploring} & 0.5310 & 0.4076 & 0.3319 & 0.2788 & 0.2490 & 0.4922 & 0.7019 \\
    \hline
    CSMLF-FT \cite{wang2019semantic} & 0.5998 & 0.4583 & 0.3869 & 0.3433 & 0.2475 & 0.5018 & 0.7555 \\
    \hline
    SVM-DCONC \cite{hoxha2021novel} & 0.7547 & 0.6711 & 0.5970 & 0.5308 & 0.3643 & 0.6746 & 2.2222 \\
    \hline
    TextGCN \cite{das2024textgcn} & 0.7680 & 0.6892 & 0.6261 & 0.5786 & 0.4009 & 0.7314 & 2.8595 \\
    \hline
    TrTr-CMR \cite{wu2024trtr} & 0.8270 & 0.6994 & 0.6002 & 0.5199 & 0.3803 & 0.7220 & 2.2728 \\
    \hline
    ResNet-LSTM~\cite{das2024unveiling} & 0.7417 & 0.6592 & 0.5925 & 0.5343 & 0.3817 & 0.6903 & 2.2032 \\
    \hline
    ConvNext-MHT~\cite{das2025good} & 0.7997 & 0.6844 & 0.6325 & 0.5694 & 0.4073 & 0.7349 & 2.4945 \\
    \hline
    JSSFF (Proposed) & \textbf{0.8402}	& \textbf{0.7738}	& \textbf{0.7196} & \textbf{0.6719}	& \textbf{0.4730}	& \textbf{0.7909}	& \textbf{3.0766} \\
    \hline
    \end{tabular}}}
\end{table}
\begin{table}[!ht]
    \centering
    \caption{Evaluation of Different Remote Sensing Image Captioning Methods on the UCM Dataset}
    \label{UCM_SEARCH}
    \resizebox{\linewidth}{!}{
    {\small
    \begin{tabular}{|c|c|c|c|c|c|c|c|}
    \hline
    METHOD & BLEU-1 & BLEU-2 & BLEU-3 & BLEU-4 & METEOR & ROUGE-L & CIDEr \\
    \hline
    R-BOW \cite{lu2017exploring} & 0.4107 & 0.2249 & 0.1452 & 0.1095 & 0.1098 & 0.3439 & 0.3071 \\
    \hline
    CSMLF-FT \cite{wang2019semantic} & 0.3671 & 0.1485 & 0.0763 & 0.0505 & 0.0944 & 0.2986 & 0.1351 \\
    \hline
    SVM-DCONC \cite{hoxha2021novel} & 0.7653 & 0.6947 & 0.6417 & 0.5942 & 0.3702 & 0.6877 & 2.9228 \\
    \hline
    TextGCN \cite{das2024textgcn} & 0.8461 & 0.7844 & 0.7386 & 0.6930 & 0.4868 & 0.8071 & 3.4077 \\
    \hline
    TrTr-CMR \cite{wu2024trtr} & 0.8156 & 0.7091 & 0.6220 & 0.5469 & 0.3978 & 0.7442 & 2.4742 \\
    \hline
    ResNet-LSTM~\cite{das2024unveiling} & 0.8001 & 0.7273 & 0.6675 & 0.6131 & 0.4084 & 0.7501 & 3.0616 \\
    \hline
    ConvNext-MHT~\cite{das2025good} & 0.8369 & 0.7712 & 0.7143 & 0.6612 & 0.4566 & 0.8119 & 3.4582 \\
    \hline
    JSSFF (Proposed) & \textbf{0.8978}	& \textbf{0.8283}	& \textbf{0.7715} & \textbf{0.7396}	& \textbf{0.5296}	& \textbf{0.8592}	& \textbf{3.8932} \\
    \hline
    \end{tabular}}}
\end{table}
\begin{table}[!ht]
    \centering
    \caption{Evaluation of Different Remote Sensing Image Captioning Methods on the RSICD Dataset}
    \label{RSICD_SEARCH}
    \resizebox{\linewidth}{!}{
    {\small
    \begin{tabular}{|c|c|c|c|c|c|c|c|}
    \hline
    METHOD & BLEU-1 & BLEU-2 & BLEU-3 & BLEU-4 & METEOR & ROUGE-L & CIDEr \\
    \hline
    R-BOW \cite{lu2017exploring} & 0.4401 & 0.2383 & 0.1514 & 0.1041 & 0.1684 & 0.3605 & 0.4667 \\
    \hline
    CSMLF-FT \cite{wang2019semantic} & 0.5106 & 0.2911 & 0.1903 & 0.1352 & 0.1693 & 0.3789 & 0.3388 \\
    \hline
    SVM-DCONC \cite{hoxha2021novel} & 0.5999 & 0.4347 & 0.3355 & 0.2689 & 0.2299 & 0.4577 & 0.6854 \\
    \hline
    TextGCN \cite{das2024textgcn} & 0.6513 & 0.4819 & 0.3747 & 0.3085 & 0.2752 & 0.4804 & 0.8266 \\
    \hline
    TrTr-CMR \cite{wu2024trtr} & 0.6201 & 0.3937 & 0.2671 & 0.1932 & 0.2399 & 0.4895 & 0.7518 \\
    \hline
    ResNet-LSTM~\cite{das2024unveiling} & 0.6407 & 0.4676 & 0.3608 & 0.2878 & 0.2574 & 0.4745 & 0.7990 \\
    \hline
    ConvNext-MHT~\cite{das2025good} & 0.6431 & 0.4665 & 0.3602 & 0.3013 & 0.2560 & 0.4945 & 0.8415 \\
    \hline
    JSSFF (Proposed) & \textbf{0.6630}	& \textbf{0.4904}	& \textbf{0.3883}	& \textbf{0.3204}	& \textbf{0.2795}	& \textbf{0.5144}	& \textbf{0.8842} \\
    \hline
    \end{tabular}}}
\end{table}

\Cref{SYDNEY_SEARCH,UCM_SEARCH,RSICD_SEARCH} presents a quantitative comparison between the proposed model (JSSFF) and several state-of-the-art RSIC models. The experimental findings confirm that the proposed JSSFF model exhibits superior performance compared to all baseline models.
\subsection{Ablation Studies}
\begin{figure*}[!ht]
  \centering
  \subfloat[\label{ablation1}]{\includegraphics[width=0.25\linewidth,keepaspectratio]{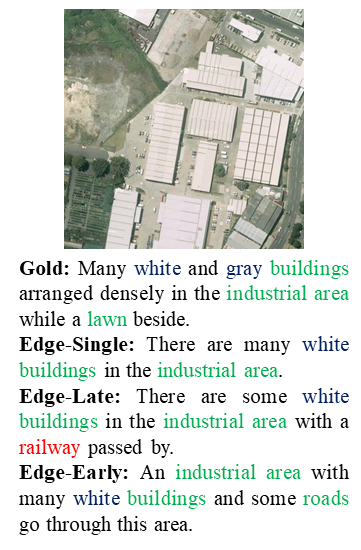}}
  \subfloat[\label{ablation2}]{\includegraphics[width=0.25\linewidth,keepaspectratio]{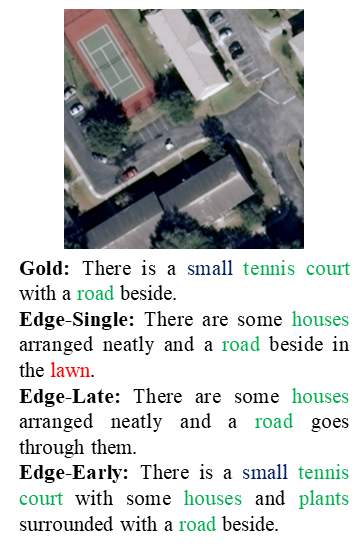}}
  \subfloat[\label{ablation3}]{\includegraphics[width=0.25\linewidth,keepaspectratio]{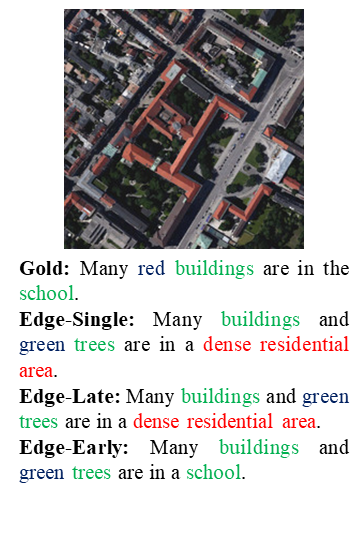}}
  \caption{Visual Examples of Ablation Study of Proposed Method}
  \label{fig:ablation}
\end{figure*}

\Cref{fig:ablation} illustrates the impact of different components in our framework. The Laplacian edge detection algorithm is used for this analysis. The conventional single-stream encoder--decoder architecture is indicated as \emph{Single}, while \emph{Early} and \emph{Late} indicate early and late fusion strategies, respectively. In addition, \emph{Edge-Early} denotes the proposed method, \emph{JSSFF}. In~\Cref{ablation1}, \emph{Edge-Single} misidentified \emph{roads} , while \emph{Edge-Late} misclassifies the image as \emph{railway}. In~\Cref{ablation2} both \emph{Edge-Single} and \emph{Edge-late} cannot detect \emph{tennis court}. In~\Cref{ablation3} both \emph{Edge-Single} and \emph{Edge-Late} misclassifies \emph{school} as \emph{dense residential area}. These results indicate that combining the original images and its edge-aware version provides more semantically accurate captions for complex remote sensing images.
\subsection{Visual Interpretation}
\begin{figure*}[!ht]
  \centering
  \subfloat[\label{example1}]{\includegraphics[width=0.25\linewidth,keepaspectratio]{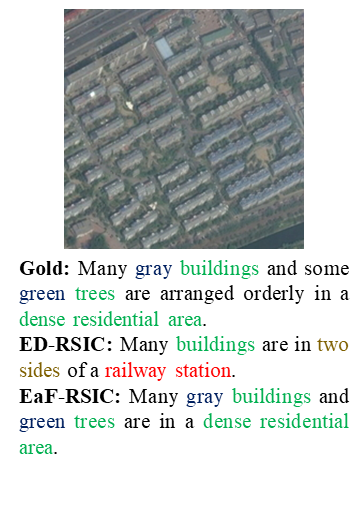}}
  \subfloat[\label{example2}]{\includegraphics[width=0.25\linewidth,keepaspectratio]{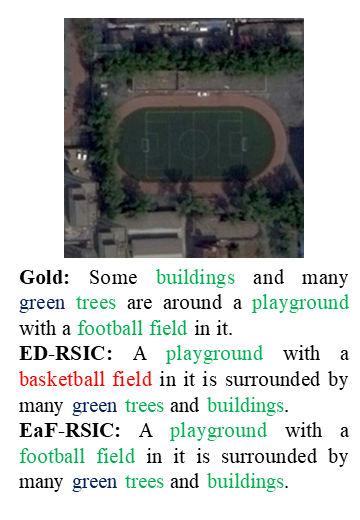}}
  \subfloat[\label{example3}]{\includegraphics[width=0.25\linewidth,keepaspectratio]{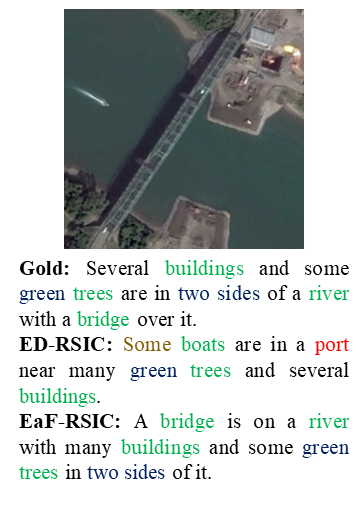}}
  
  \subfloat[\label{example4}]{\includegraphics[width=0.25\linewidth,keepaspectratio]{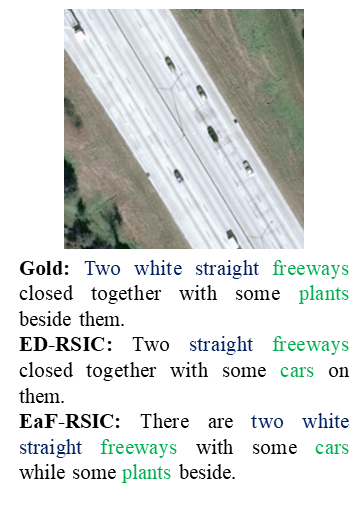}}
  \subfloat[\label{example5}]{\includegraphics[width=0.25\linewidth,keepaspectratio]{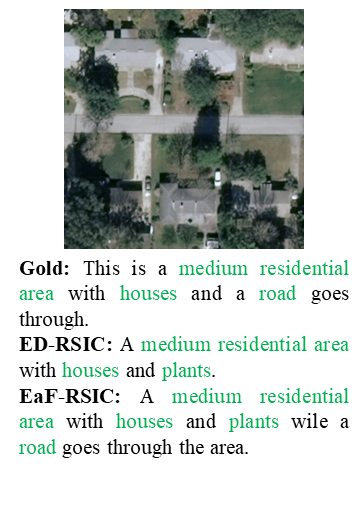}}
  \subfloat[\label{example6}]{\includegraphics[width=0.25\linewidth,keepaspectratio]{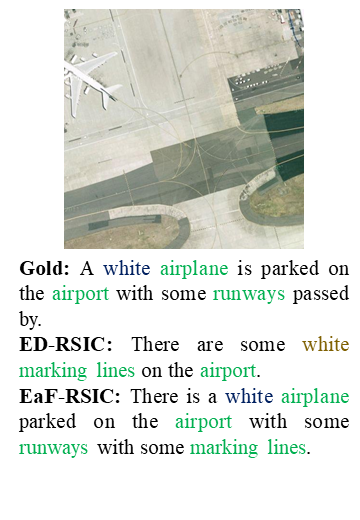}}

  \subfloat[\label{example7}]{\includegraphics[width=0.25\linewidth,keepaspectratio]{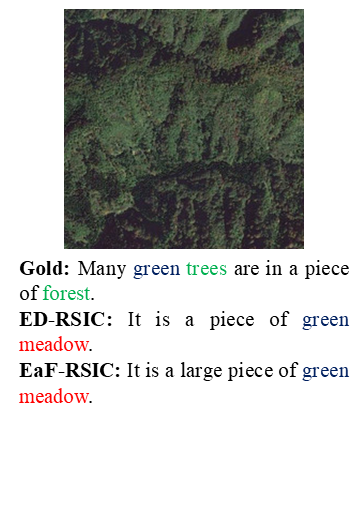}}
  \subfloat[\label{example8}]{\includegraphics[width=0.25\linewidth,keepaspectratio]{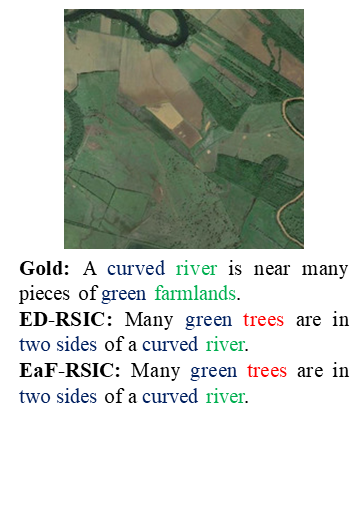}}
  \subfloat[\label{example9}]{\includegraphics[width=0.25\linewidth,keepaspectratio]{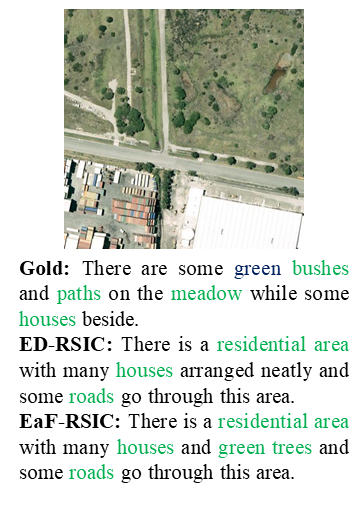}}
   \caption{Visual Examples of Various RSIC Methods}
   \label{fig:visual}
\end{figure*}

\Cref{fig:visual} illustrates the visual comparison of the proposed framework (JSSFF) with reference captions (Gold) and the traditional encoder–decoder framework (ED-RSIC). These examples illustrate instances where the proposed method generates captions that are semantically richer. 

In the first set of examples (\Cref{example1,example2,example3}), the ED-RSIC model produces incorrect or misleading captions, whereas the JSSFF generates accurate and contextually relevant descriptions. For example, the ED-RSIC model misclassifies \emph{dense residential area} as \emph{railway station} (\Cref{example1}); mislabels \emph{football field} as \emph{basketball field} (\Cref{example2}); and in \Cref{example3}, incorrectly describes \emph{river} as \emph{port}, fails to detect the major object \emph{bridge}, and incorrectly identifies \emph{a boat} as \emph{some boats}.

The second set of examples (\Cref{example4,example5,example6}) shows that the ED-RSIC model incorporates additional visual and semantic details that improve the overall quality of the generated captions. For example, the JSSFF model successfully detects \emph{plants} beside \emph{freeways} (\Cref{example4}); identifies \emph{road} in \emph{medium residential area} image (\Cref{example5}); and in \Cref{example6}, recognizes both the primary object \emph{airport} and the secondary object \emph{runways}, while the ED-RSIC model misrepresents the color of \emph{marking} line as \emph{white}, although it is actually \emph{yellow}.

The third set of examples (\Cref{example7,example8,example9}) presents failure cases that are commonly observed in both our model and the baseline. For example, both models misclassify \emph{forest} image as \emph{meadow} (\Cref{example7}); mislabel \emph{farmlands} as\emph{meadow} (\Cref{example8}); and misidentify the core object \emph{meadow} while incorrectly classifying the image as \emph{residential area} (\Cref{example9}).
\subsection{Subjective Evaluation}
\label{sec:subjective}
Unlike classification tasks, image captioning can have multiple valid descriptions for the same image. Hence, evaluating generated captions solely against the five reference captions available in each of the three datasets is not sufficient \cite{das2024unveiling,lu2017exploring}. To overcome this limitation, we performed a qualitative evaluation with a human annotator\footnote{A domain expert with several years of experience in RSIC.} This human-centered assessment offers deeper insights into the semantic quality of the captions. In our evaluation, captions were categorized into three labels: 
\begin{enumerate}
    \item\textbf{Related:} The caption provides an accurate textual description of the image along with its class label.
    \item\textbf{Partially Related:} The caption includes the image description and class label, but contains minor errors.
    \item\textbf{Unrelated:} The caption is inconsistent with the image content and misclassifies the image.
\end{enumerate}
\begin{table}[!ht]
\centering
\caption{Subjective Evaluation of the Baseline (ED-RSIC) and Proposed (JSSFF) method (in \%)}
\label{subjective}
\resizebox{0.75\linewidth}{!}{%
\begin{tabular}{|c|c|c|c|c|}
\hline
\textbf{Dataset} & \textbf{Method} & \textbf{Related} & \textbf{Partially Related} & \textbf{Unrelated} \\
\hline

\multirow{2}{*}{SYDNEY} 
& ED-RSIC   & 87.93 & 6.90 & 5.17 \\ \cline{2-5}
& JSSFF  & 91.38 & 5.17 & 3.45 \\
\hline

\multirow{2}{*}{UCM}
& ED-RSIC   & 89.52 & 5.72 & 4.76 \\ \cline{2-5}
& JSSFF  & 92.86 & 3.81 & 3.33 \\
\hline

\multirow{2}{*}{RSICD}
& ED-RSIC   & 86.92 & 7.23 & 5.85 \\ \cline{2-5}
& JSSFF  & 87.65 & 7.50 & 4.85 \\
\hline

\end{tabular}%
}
\end{table}

\Cref{subjective} presents the results of a subjective evaluation of different models, without relying on any quantitative metrics. Here we compare our model (JSSFF) with the conventional encoder--decoder model using the ConvNext encoder (ED-RSIC). The values denote the percentage of test images judged as \emph{Related}, \emph{Partially Related}, or \emph{Unrelated}, based on the perception of a human evaluator. These results clearly indicate that our model consistently outperforms the conventional encoder–decoder model in visual evaluation.
\subsection{Error Analysis}
Several distinct error patterns were identified in the generated captions~\cite{das2024unveiling,das2024textgcn}. The primary source of error is \emph{misclassification}, which can be classified into two categories. The first type is \emph{major object misclassification}, which occurs when the primary object in an image is incorrectly identified~\Cref{ablation1,ablation3,example7,example8}. The second type, \emph{minor object misclassification}, arises when the main object is correctly recognized, but secondary objects are mislabeled (\Cref{example2}). Another prevalent issue is \emph{semantic similarity confusion}, where classes with high visual or contextual resemblance are confused (\Cref{example3}). Another major error is \emph{missing core object}, which occurs when the model does not detect a key element within the scene (\Cref{ablation2}). Additional minor errors include \emph{missing minor objects}, often caused by the model's bias towards the central image region (\Cref{example4,example5}). Another recurring problem involves \emph{misidentifying complex spatial patterns with multiple clusters} (\Cref{example9}). Occasional errors also include \emph{counting errors} (\Cref{example3}), \emph{attribute errors} (\Cref{example6}), and \emph{sentence incompletion errors}, although these affect only a small number of test images.

From the analysis of visual examples (\Cref{example1,example2,example3,example4,example5,example6,example7,example8,example9}), we observed that edge-aware images help the model capture local visual features more effectively. As a result, images with fine structural details (such as \emph{buildings}, \emph{roads}, and \emph{airplanes}) benefit more from this approach. In contrast, images with homogeneous or smooth regions such as (\emph{forest} and \emph{farmland}) show only minor improvement because edge detection adds little information in areas with uniform texture or shape~\cite{das2025novel}.
\section{Conclusions}
\label{sec:conclusion}
Encoder--decoder frameworks have significantly improved RSIC systems but still suffer from inaccurate object detection and finding relationships among them. However, the information extracted through an encoder is beneficial for the decoder in generating meaningful descriptions of the image. We have proposed an edge-aware early fusion technique to address this. Apart from using the original image, we have also incorporated its edge-aware version to extract richer visual cues and structural details from complex scenes. In addition, early fusion of both the original and edge-aware image information can help the model better capture object boundaries and spatial relationships. In addition, while decoding the captions from the model, we used CBBS to improve the generated caption, which preserves both the visual and the textual properties.Our proposed model shows consistent improvement in RSIC, supported by extensive experimental results across all benchmark datasets. Especially, objects with sharp boundaries can be efficiently detected in the caption. In our future work, we will explore advanced yet efficient fusion and decoding strategies to further improve caption accuracy and adaptability.
\bibliographystyle{ieeetr} 
\bibliography{MyBib}

@article{hoxha2021novel,
  title={A novel SVM-based decoder for remote sensing image captioning},
  author={Hoxha, Genc and Melgani, Farid},
  journal={IEEE Transactions on Geoscience and Remote Sensing},
  volume={60},
  pages={1--14},
  year={2021},
  publisher={IEEE}
}

@inproceedings{qu2016deep,
  title={Deep semantic understanding of high resolution remote sensing image},
  author={Qu, Bo and Li, Xuelong and Tao, Dacheng and Lu, Xiaoqiang},
  booktitle={2016 International conference on computer, information and telecommunication systems (Cits)},
  pages={1--5},
  year={2016},
  organization={IEEE}
}

@article{wang2019semantic,
  title={Semantic descriptions of high-resolution remote sensing images},
  author={Wang, Binqiang and Lu, Xiaoqiang and Zheng, Xiangtao and Li, Xuelong},
  journal={IEEE Geoscience and Remote Sensing Letters},
  volume={16},
  number={8},
  pages={1274--1278},
  year={2019},
  publisher={IEEE}
}

@inproceedings{hoxha2020new,
  title={A new CNN-RNN framework for remote sensing image captioning},
  author={Hoxha, Genc and Melgani, Farid and Slaghenauffi, Jacopo},
  booktitle={2020 Mediterranean and Middle-East Geoscience and Remote Sensing Symposium (M2GARSS)},
  pages={1--4},
  year={2020},
  organization={IEEE}
}

@article{lu2017exploring,
  title={Exploring models and data for remote sensing image caption generation},
  author={Lu, Xiaoqiang and Wang, Binqiang and Zheng, Xiangtao and Li, Xuelong},
  journal={IEEE Transactions on Geoscience and Remote Sensing},
  volume={56},
  number={4},
  pages={2183--2195},
  year={2017},
  publisher={IEEE}
}

@article{ye2022joint,
  title={A joint-training two-stage method for remote sensing image captioning},
  author={Ye, Xiutiao and Wang, Shuang and Gu, Yu and Wang, Jihui and Wang, Ruixuan and Hou, Biao and Giunchiglia, Fausto and Jiao, Licheng},
  journal={IEEE Transactions on Geoscience and Remote Sensing},
  volume={60},
  pages={1--16},
  year={2022},
  publisher={IEEE}
}

@article{li2020multi,
  title={A multi-level attention model for remote sensing image captions},
  author={Li, Yangyang and Fang, Shuangkang and Jiao, Licheng and Liu, Ruijiao and Shang, Ronghua},
  journal={Remote Sensing},
  volume={12},
  number={6},
  pages={939},
  year={2020},
  publisher={MDPI}
}

@article{ren2022mask,
  title={A mask-guided transformer network with topic token for remote sensing image captioning},
  author={Ren, Zihao and Gou, Shuiping and Guo, Zhang and Mao, Shasha and Li, Ruimin},
  journal={Remote Sensing},
  volume={14},
  number={12},
  pages={2939},
  year={2022},
  publisher={MDPI}
}

@inproceedings{papineni-etal-2002-bleu,
    title = "{B}leu: a Method for Automatic Evaluation of Machine Translation",
    author = "Papineni, Kishore  and
      Roukos, Salim  and
      Ward, Todd  and
      Zhu, Wei-Jing",
    booktitle = "Proceedings of the 40th Annual Meeting of the Association for Computational Linguistics",
    month = jul,
    year = "2002",
    address = "Philadelphia, Pennsylvania, USA",
    publisher = "Association for Computational Linguistics",
    url = "https://aclanthology.org/P02-1040",
    doi = "10.3115/1073083.1073135",
    pages = "311--318",
}

@inproceedings{lavie-agarwal-2007-meteor,
    title = "{METEOR}: An Automatic Metric for {MT} Evaluation with High Levels of Correlation with Human Judgments",
    author = "Lavie, Alon  and
      Agarwal, Abhaya",
    booktitle = "Proceedings of the Second Workshop on Statistical Machine Translation",
    month = jun,
    year = "2007",
    address = "Prague, Czech Republic",
    publisher = "Association for Computational Linguistics",
    url = "https://aclanthology.org/W07-0734",
    pages = "228--231",
}

@inproceedings{vedantam2015cider,
  title={Cider: Consensus-based image description evaluation},
  author={Vedantam, Ramakrishna and Lawrence Zitnick, C and Parikh, Devi},
  booktitle={Proceedings of the IEEE conference on computer vision and pattern recognition},
  pages={4566--4575},
  year={2015}
}

@inproceedings{lin-2004-rouge,
    title = "{ROUGE}: A Package for Automatic Evaluation of Summaries",
    author = "Lin, Chin-Yew",
    booktitle = "Text Summarization Branches Out",
    month = jul,
    year = "2004",
    address = "Barcelona, Spain",
    publisher = "Association for Computational Linguistics",
    url = "https://aclanthology.org/W04-1013",
    pages = "74--81",
}

@inproceedings{das2024unveiling,
  title={Unveiling the Power of Convolutional Neural Networks: A Comprehensive Study on Remote Sensing Image Captioning and Encoder Selection},
  author={Das, Swadhin and Khandelwal, Akshat and Sharma, Raksha},
  booktitle={2024 International Joint Conference on Neural Networks (IJCNN)},
  pages={1--8},
  year={2024},
  organization={IEEE}
}

@article{das2024textgcn,
  title={A TextGCN-Based Decoding Approach for Improving Remote Sensing Image Captioning},
  author={Das, Swadhin and Sharma, Raksha},
  journal={IEEE Geoscience and Remote Sensing Letters},
  year={2024},
  publisher={IEEE}
}

@article{wu2024trtr,
  title={TrTr-CMR: Cross-Modal Reasoning Dual Transformer for Remote Sensing Image Captioning},
  author={Wu, Yinan and Li, Lingling and Jiao, Licheng and Liu, Fang and Liu, Xu and Yang, Shuyuan},
  journal={IEEE Transactions on Geoscience and Remote Sensing},
  year={2024},
  publisher={IEEE}
}

@inproceedings{liu2022convnet,
  title={A convnet for the 2020s},
  author={Liu, Zhuang and Mao, Hanzi and Wu, Chao-Yuan and Feichtenhofer, Christoph and Darrell, Trevor and Xie, Saining},
  booktitle={Proceedings of the IEEE/CVF conference on computer vision and pattern recognition},
  pages={11976--11986},
  year={2022}
}

@article{das2025good,
  title={Good Representation, Better Explanation: Role of Convolutional Neural Networks in Transformer-Based Remote Sensing Image Captioning},
  author={Das, Swadhin and Gupta, Saarthak and Kumar, Kamal and Sharma, Raksha},
  journal={arXiv preprint arXiv:2502.16095},
  year={2025}
}

@inproceedings{sutskever2014seq2seq,
  title={Sequence to Sequence Learning with Neural Networks},
  author={Sutskever, Ilya and Vinyals, Oriol and Le, Quoc V},
  booktitle={Advances in neural information processing systems},
  pages={3104--3112},
  year={2014}
}

@article{bahdanau2014neural,
  title={Neural Machine Translation by Jointly Learning to Align and Translate},
  author={Bahdanau, Dzmitry and Cho, Kyunghyun and Bengio, Yoshua},
  journal={arXiv preprint arXiv:1409.0473},
  year={2014}
}

@article{das2025fe,
  title={FE-LWS: Refined Image-Text Representations via Decoder Stacking and Fused Encodings for Remote Sensing Image Captioning},
  author={Das, Swadhin and Sharma, Raksha},
  journal={arXiv preprint arXiv:2502.09282},
  year={2025}
}

@article{das2025novel,
  title={A Novel Lightweight Transformer with Edge-Aware Fusion for Remote Sensing Image Captioning},
  author={Das, Swadhin and Mundra, Divyansh and Dayal, Priyanshu and Sharma, Raksha},
  journal={arXiv preprint arXiv:2506.09429},
  year={2025}
}

@article{canny1986computational,
  title={A computational approach to edge detection},
  author={Canny, John},
  journal={IEEE Transactions on pattern analysis and machine intelligence},
  number={6},
  pages={679--698},
  year={1986},
  publisher={Ieee}
}

@article{sobel19683x3,
  title={A 3x3 isotropic gradient operator for image processing},
  author={Sobel, Irwin and Feldman, Gary and others},
  journal={a talk at the Stanford Artificial Project in},
  volume={1968},
  pages={271--272},
  year={1968}
}

@article{marr1980theory,
  title={Theory of edge detection},
  author={Marr, David and Hildreth, Ellen},
  journal={Proceedings of the Royal Society of London. Series B. Biological Sciences},
  volume={207},
  number={1167},
  pages={187--217},
  year={1980},
  publisher={The Royal Society London}
}

@inproceedings{zhang2019multi,
  title={Multi-scale cropping mechanism for remote sensing image captioning},
  author={Zhang, Xueting and Wang, Qi and Chen, Shangdong and Li, Xuelong},
  booktitle={IGARSS 2019-2019 IEEE International Geoscience and Remote Sensing Symposium},
  pages={10039--10042},
  year={2019},
  organization={IEEE}
}

@inproceedings{das2020pso,
  title={Correcting Low Illumination Images Using PSO-Based Gamma Correction and Image Classifying Method},
  author={Das, Swadhin and Roy, Manali and Mukhopadhyay, Susanta},
  booktitle={International Conference on Computer Vision and Image Processing},
  pages={433--444},
  year={2020},
  organization={Springer}
}

@inproceedings{das2020correcting,
  title={Correcting Low-Illumination Images Using Multi-Scale Fusion in a Pyramidal Framework},
  author={Das, Swadhin and Roy, Manali and Mukhopadhyay, Susanta},
  booktitle={2020 International Conference on Wireless Communications Signal Processing and Networking (WiSPNET)},
  pages={126--129},
  year={2020},
  organization={IEEE}
}

@article{hoxha2023improving,
  title={Improving image captioning systems with postprocessing strategies},
  author={Hoxha, Genc and Scuccato, Giacomo and Melgani, Farid},
  journal={IEEE Transactions on Geoscience and Remote Sensing},
  volume={61},
  pages={1--13},
  year={2023},
  publisher={IEEE}
}

@article{wang2022multi,
  title={Multi-label semantic feature fusion for remote sensing image captioning},
  author={Wang, Shuang and Ye, Xiutiao and Gu, Yu and Wang, Jihui and Meng, Yun and Tian, Jingxian and Hou, Biao and Jiao, Licheng},
  journal={ISPRS Journal of Photogrammetry and Remote Sensing},
  volume={184},
  pages={1--18},
  year={2022},
  publisher={Elsevier}
}

@inproceedings{chen2018show,
  title={Show, Observe and Tell: Attribute-driven Attention Model for Image Captioning.},
  author={Chen, Hui and Ding, Guiguang and Lin, Zijia and Zhao, Sicheng and Han, Jungong},
  booktitle={IJCAI},
  pages={606--612},
  year={2018}
}

@article{zia2022transforming,
  title={Transforming remote sensing images to textual descriptions},
  author={Zia, Usman and Riaz, M Mohsin and Ghafoor, Abdul},
  journal={International Journal of Applied Earth Observation and Geoinformation},
  volume={108},
  pages={102741},
  year={2022},
  publisher={Elsevier}
}

@inproceedings{xu2017image,
  title={Image captioning with deep LSTM based on sequential residual},
  author={Xu, Kaisheng and Wang, Hanli and Tang, Pengjie},
  booktitle={2017 IEEE International Conference on Multimedia and Expo (ICME)},
  pages={361--366},
  year={2017},
  organization={IEEE}
}

@inproceedings{sairam2021image,
  title={Image Captioning using CNN and LSTM},
  author={Sairam, Gourishetty and Mandha, Mounika and Prashanth, Penjarla and Swetha, Polisetty},
  booktitle={4th Smart cities symposium (SCS 2021)},
  volume={2021},
  pages={274--277},
  year={2021},
  organization={IET}
}
\end{document}